\documentclass{article}

\usepackage[preprint]{neurips_2025}

\usepackage[utf8]{inputenc} 
\usepackage[T1]{fontenc}    
\usepackage{hyperref}       
\usepackage{url}            
\usepackage{booktabs}       
\usepackage{amsfonts}       
\usepackage{nicefrac}       
\usepackage{microtype}      
\usepackage{xcolor}         

\title{Log-Normal Multiplicative Dynamics for \\ Stable Low-Precision Training of Large Networks}
\usepackage{amsmath}
\usepackage{algorithm}
\usepackage{algorithmic}
\usepackage{graphicx}

\newcommand{\vparam}{\vtheta}
\newcommand{\param}{\theta}

\newcommand{\dkl}[3]{\mathbb{D}_{\text{KL}}^{#1}(#2 \, \|\, #3)}

\newcommand\cut[1]{}

\newcommand{\squishlist}{
   \begin{list}{$\bullet$}
    { \setlength{\itemsep}{0pt}      \setlength{\parsep}{3pt}
      \setlength{\topsep}{3pt}       \setlength{\partopsep}{0pt}
      \setlength{\leftmargin}{1.5em} \setlength{\labelwidth}{1em}
      \setlength{\labelsep}{0.5em} } }

\newcommand{\squishlisttwo}{
   \begin{list}{$\bullet$}
    { \setlength{\itemsep}{0pt}    \setlength{\parsep}{0pt}
      \setlength{\topsep}{0pt}     \setlength{\partopsep}{0pt}
      \setlength{\leftmargin}{2em} \setlength{\labelwidth}{1.5em}
      \setlength{\labelsep}{0.5em} } }

\newcommand{\squishend}{
    \end{list}  }

{}
{}
{}
{}

\newcommand{\myvec}[1]{\mbox{$\mathbf{#1}$}}
\newcommand{\myvecsym}[1]{\mbox{$\boldsymbol{#1}$}}

\newcommand{\vmu}{\mbox{$\myvecsym{\mu}$}}
\newcommand{\vnu}{\mbox{$\myvecsym{\nu}$}}

\newcommand{\vtheta}{\mbox{$\myvecsym{\theta}$}}

\newcommand{\vg}{\mbox{$\myvec{g}$}}

\newcommand{\vm}{\mbox{$\myvec{m}$}}

\newcommand{\vr}{\mbox{$\myvec{r}$}}

\newcommand{\vA}{\mbox{$\myvec{A}$}}

\newcommand{\vI}{\mbox{$\myvec{I}$}}

\newcommand{\be}{\begin{equation}}
\newcommand{\ee}{\end{equation}}
\newcommand{\bea}{\begin{eqnarray}}
\newcommand{\eea}{\end{eqnarray}}
\newcommand{\beaa}{\begin{eqnarray*}}
\newcommand{\eeaa}{\end{eqnarray*}}

\newcommand{\blockcomment}[1]{}

\newcommand*\iftodonotes{\if@todonotes@disabled\expandafter\@secondoftwo\else\expandafter\@firstoftwo\fi}  
\makeatother

\usepackage{booktabs}
\usepackage{multirow}
\usepackage{colortbl}
\definecolor{LightGray}{gray}{0.90}

\usepackage{xurl}

\usepackage{inconsolata}
\definecolor{codebg}{gray}{0.95}

\definecolor{codebg}{rgb}{0.95,0.95,0.95}
\definecolor{addcolor}{rgb}{0,0.6,0}
\definecolor{delcolor}{rgb}{0.8,0,0}
\usepackage{listings}
\usepackage{wrapfig} 

\usepackage{soul} 

\author{%
  Keigo Nishida
  \thanks{Corresponding author.}
  \hspace{.02in} 
  \thanks{RIKEN Center for Biosystems Dynamics Research, Kobe, Japan.}
  \hspace{.02in}
  \thanks{RIKEN Center for AI Project, Tokyo, Japan.}\\
  \texttt{keigo.nishida.jg@riken.jp}\\
  \And
  Eren Mehmet Kıral
  $^\ddagger$
  \thanks{Keio University, Yokohama, Japan.} \\
  \texttt{eren.kiral@riken.jp}\\
  \And 
  Kenichi Bannai
  $^\ddagger$
  $^\S$\\
  \texttt{kenichi.bannai@riken.jp}\\
  \And
  Mohammad Emtiyaz Khan
  $^\ddagger$\\
  \texttt{emtiyaz.khan@riken.jp}\\
  \And
  Thomas M\"ollenhoff
  $^\ddagger$\\
  \texttt{thomas.moellenhoff@riken.jp}\\
}

\begin{document}

\maketitle

\begin{abstract}
Studies in neuroscience have shown that biological synapses follow a log-normal distribution whose transitioning can be explained by noisy multiplicative dynamics. Biological networks can function stably even under dynamically fluctuating conditions arising due to unreliable synaptic transmissions. Here we ask: Is it possible to design similar multiplicative training in artificial neural networks? To answer this question, we derive a Bayesian learning rule that assumes log-normal posterior distributions over weights which gives rise to a new Log-Normal Multiplicative Dynamics (LMD) algorithm. The algorithm uses multiplicative updates with both noise and regularization applied multiplicatively. The method is as easy to implement as Adam and only requires one additional vector to store. Our results show that LMD achieves stable and accurate training-from-scratch under low-precision forward operations for Vision Transformer and GPT-2. These results suggest that multiplicative dynamics, a biological feature, may enable stable low-precision inference and learning on future energy-efficient hardware. Code is available at \url{https://github.com/team-approx-bayes/lmd}
\end{abstract}

\section{Introduction}
\label{introduction}
Biological synapses exhibit continuous dynamic activity accompanied by noise~\citep{Choquet2013}. The distribution of the synaptic spine size has been observed to follow a log-normal distribution, which is thought to arise from noisy multiplicative dynamics~\citep{Loewenstein2011}. In addition to exhibiting activity-dependent plasticity, synaptic spines also show spontaneous fluctuations, with their dynamics considered to be approximately proportional to the spine size~\citep{Loewenstein2011, Kasai2021}.
Such dynamics are often regarded as a cause of unreliable or uncertain information transmission, but their importance in neural computation is being increasingly recognized~\citep{Seung2003,Bartol2015,Kappel2015,Aitchison2021}.

To exploit the dynamics of biological synapses for artificial neural networks (ANNs), both the weight updates and the noise injection for the spontaneous fluctuations need to be multiplicative. However, contemporary ANN training employ only one of these two mechanisms. ~\citet{Bernstein2020} addressed the question of how the brain can stably learn despite the unreliability of biological synapses by adopting multiplicative weight updates (MWU)~\citep{Arora2012} for ANNs. However, spontaneous fluctuations caused by noise were not addressed. In contrast, Bayesian neural networks assume a probability distribution over their weights and sample from this distribution~\citep{Graves2011,Blundell2015,Khan2018}. However, because the sampling noise is not inherently proportional to the weight magnitude, it is insufficient to reproduce the fluctuations observed in biological synapses. Previous works use multiplicative noise proportional to the weight norm, but still use additive gradient updates~\citep{Yeming2018,bisla2022low,Trung2024}. 

Multiplicative fluctuation is highly compatible with data formats that emphasize dynamic range with limited bit width, making it promising for application to energy-efficient dedicated hardware required by low-precision data formats for both inference and training~\citep{Bernstein2020,Zhao2022,Haghi2024}.
This characteristic may also help improve computational throughput using the Microscaling (MX) data format~\citep{Rouhani2023OCP}, which has been considered for various LLM deployments~\citep{Bita2023,Verrilli2024,Tseng2025,Ramani2025}. However, multiplicative updates tend to cause exponential weight growth~\citep[Section~8]{Bernstein2020}, and their use in training large-scale neural networks from scratch has not been successful.
Stabilizing learning under multiplicative updates requires new regularization techniques. 

\begin{figure}[!t]
  \centering
  \includegraphics[width=0.49\textwidth]{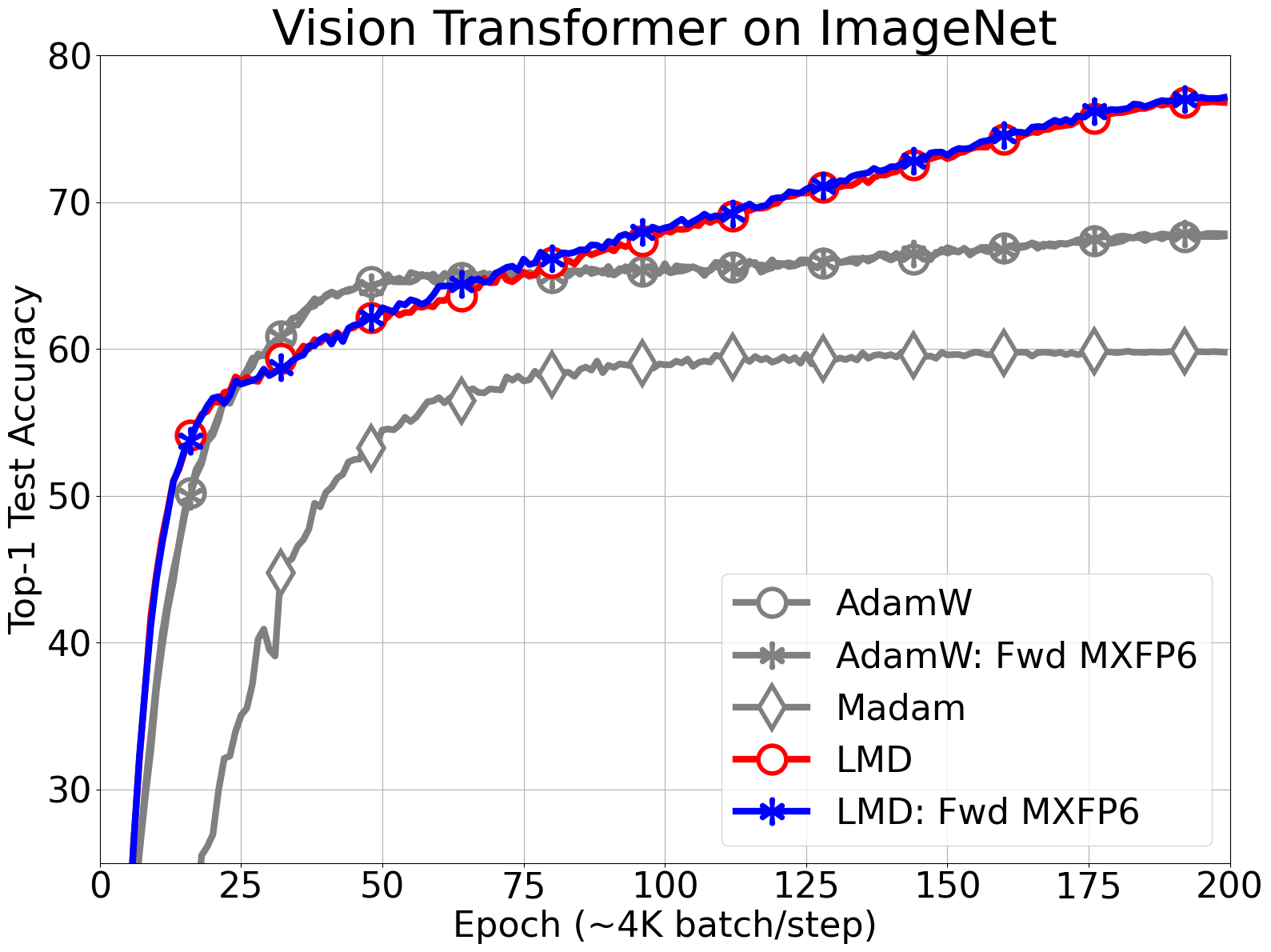}
  \includegraphics[width=0.49\textwidth]{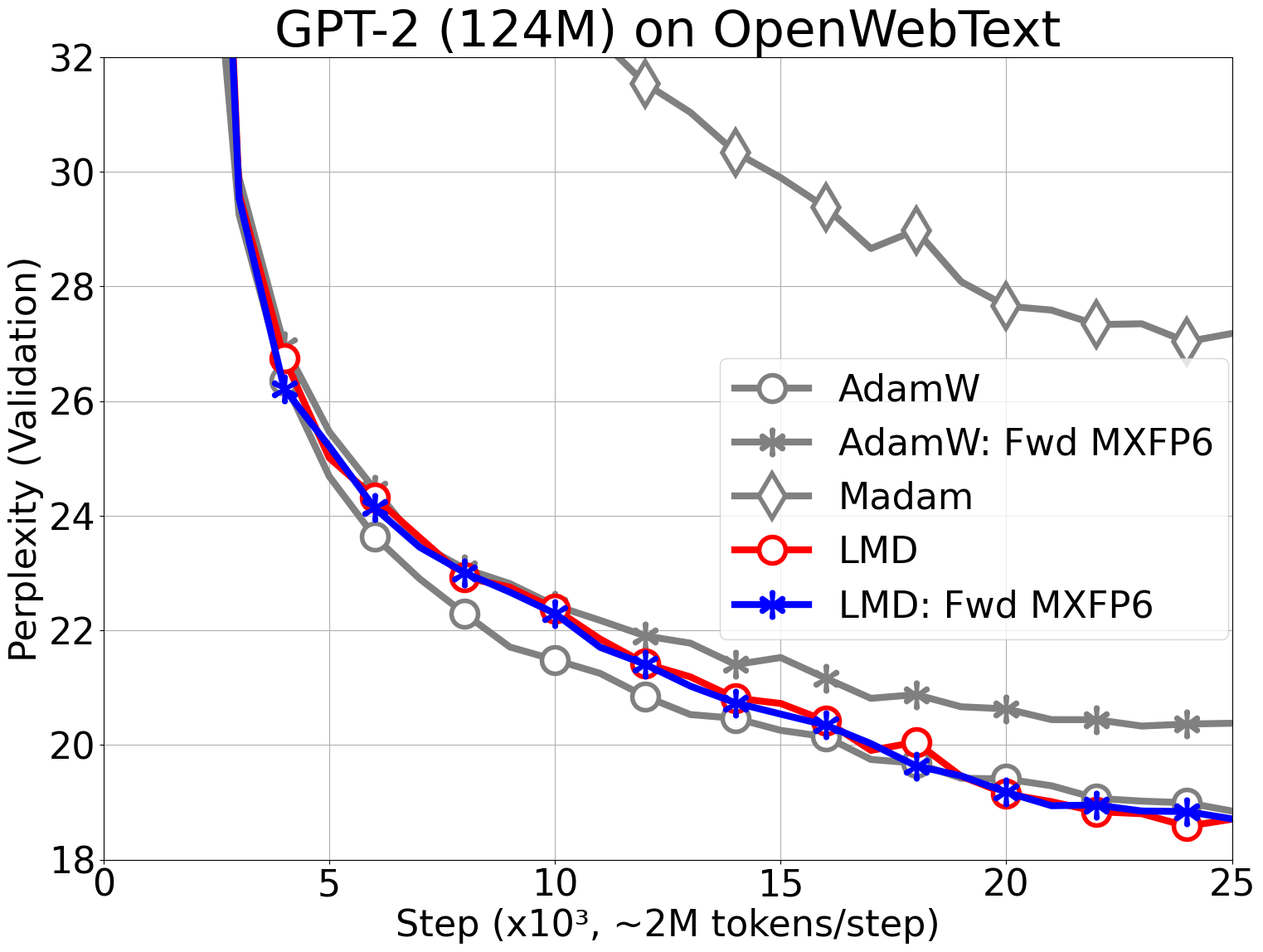}
  \caption{These two plots demonstrate that LMD enables full-scratch learning of Vision Transformer and GPT-2. Even when the forward pass is performed using low-precision formats based on the Microscaling data format (MXFP6), no performance degradation occurs. AdamW and Madam are unable to sufficiently train ViT. On GPT-2, AdamW shows perplexity similar to LMD, but we can observe that the optimizer can not stably learn using low-precision forward-passes.
  For GPT-2 training, LMD and Madam use a sequence length of 4096 and a batch size of 16. Note that AdamW uses a sequence length of 1024 and a batch size of 64. However, the number of tokens per step is the same. All training, except for the forward passes in MXFP6, was performed using bfloat16.
  }
  \label{fig:fig1}
\end{figure}

In this paper, we propose the Log-Normal Multiplicative Dynamics (LMD) algorithm. LMD is derived from the Lie-Group Bayesian Learning Rule (Lie-Group BLR) with the multiplicative group of positive reals and log-normal weight noise distribution~\citep{Kiral2023}. 
LMD uses multiplicative weight updates with both noise and regularization applied multiplicatively, using which we can train large ANNs from scratch successfully, see Figure~\ref{fig:fig1}. 

\begin{figure}[!t]
  \centering
  \includegraphics[width=0.49\textwidth]{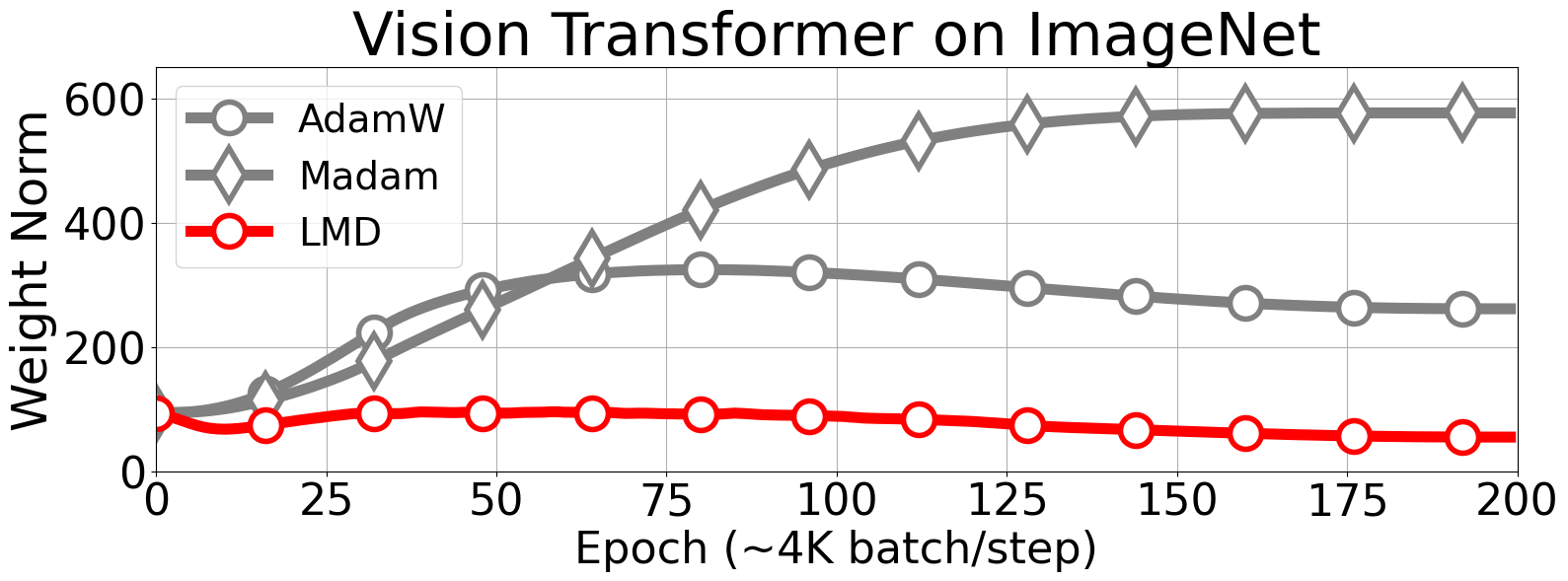}
  \includegraphics[width=0.49\textwidth]{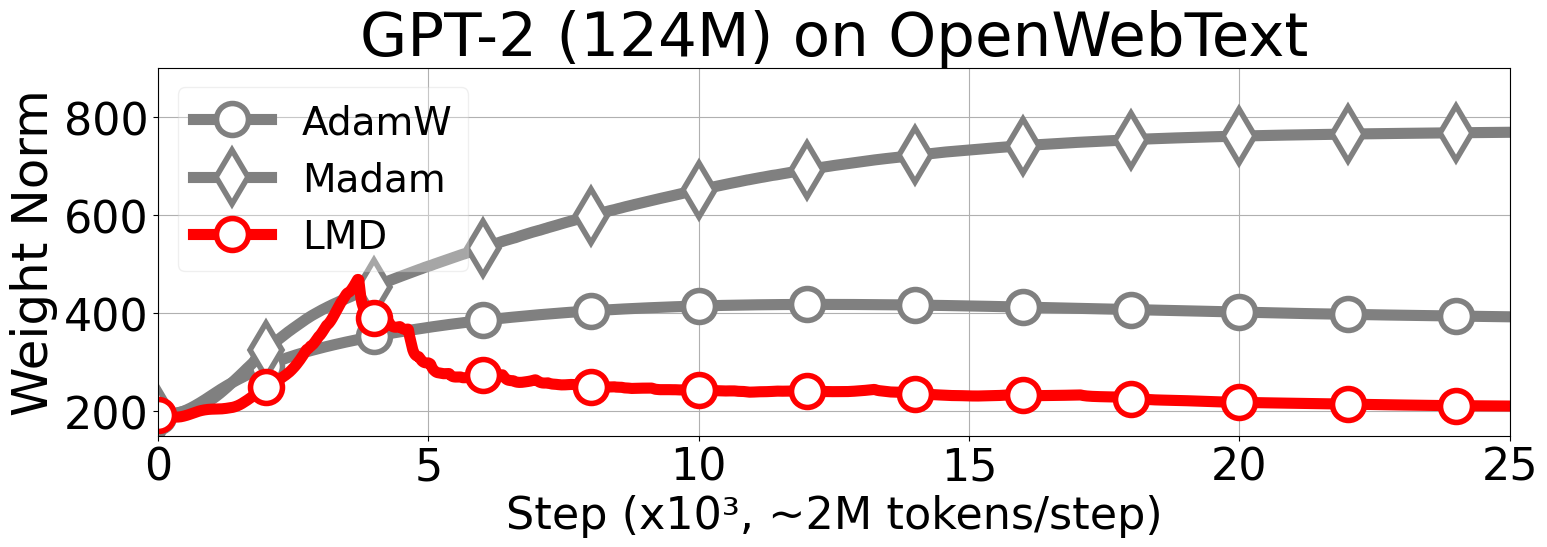}
  \caption{These two plots show the dynamics of the \(\ell_{2}\) norm of weights during training of the models shown in Figure 1 when only using bfloat16. Madam takes a large norm, while LMD converges to a norm close to the initial value, clearly suppressing excessive weight increase.}
  \label{fig:fig2}
\end{figure}

Our contributions are as follows:
\begin{enumerate}
  \item We achieve the first successful training from scratch of Vision Transformer (ViT)~\citep{Alexey2021} and GPT-2~\citep{Radford2019} using multiplicative weight updates. Using low-precision emulation, we demonstrate that LMD maintains full model performance even when forward passes are performed in the low-precision MX data format (Figure \ref{fig:fig1}). 
  \item We show that LMD effectively suppresses the excessive weight growth that has been an issue for standard MWU methods (Figure \ref{fig:fig2}). Through ablation studies, we verify that multiplicative weight decay strongly suppresses weight growth and delivers superior regularization compared to conventional additive weight decay (Figure \ref{fig:fig3}).
  \item We show that multiplicative noise injection is important for stabilizing the weights when using low-precision MX data format for forward passes (Figure \ref{fig:fig4}). (Backward passes use bfloat16~\citep{Kalamkar2019} rather than the MX data format.)
\end{enumerate}

\begin{table}[t]
  \caption{ViT and GPT-2 optimizer comparison; mean ± standard deviation (n = 3 independent runs).}
  \label{tab:results}
  \centering
  \begin{tabular}{lccc c ccc}
  \toprule
    & \multicolumn{2}{c}{ViT on ImageNet} & \multicolumn{3}{c}{GPT-2 on OpenWebText} \\
   \cmidrule(lr){2-3} \cmidrule(lr){4-6}
    Method & Test Acc. (\%) &  Weight Norm
           & (seq-len) & Val. Loss  & Weight Norm \\
    \midrule
    \rowcolor{LightGray}
    AdamW               &  $68.11_{\pm  0.38}$   &   $260.7_{\pm  0.5}$   & 1024 & $2.937_{\pm  0.001}$ & $392.7_{\pm  0.4}$ \\
    \rowcolor{LightGray}
                        &    &             & 4096 & $4.790_{\pm 2.017}$ & $931.7_{\pm  70.5}$ \\

   ~~w/ Fwd MXFP6      &  $67.99_{\pm  0.27}$  &     $261.5_{\pm  0.4}$  & 1024 & $3.015_{\pm  0.000}$  & $425.0_{\pm  0.4}$ \\

    \rowcolor{LightGray}
    Madam               &  $60.14 _{\pm  0.31}$  &    $577.3 _{\pm  0.9}$  & 4096 & $3.303_{\pm  0.003}$  & $765.1_{\pm  3.4}$ \\

    LMD                 &  $ 77.06 _{\pm  0.08}$   &  $55.2 _{\pm  0.1} $ & 1024 & $2.961_{\pm  0.002}$  & $217.2_{\pm  4.5} $\\
                        &     &            & 4096 & $2.925_{\pm  0.006}$  & $212.9_{\pm  2.1}$ \\

    \rowcolor{LightGray}
    ~~w/ Fwd MXFP6        &$ \bf 77.15_{\pm  0.08} $ &   $ 55.5_{\pm  0.3}$   & 4096 & $ \bf 2.927_{\pm  0.002}$  & $204.1_{\pm  11.2}$ \\
    \bottomrule
  \end{tabular}
\end{table}

\section{Background and Related Work}
\subsection{Multiplicative Weight Updates}
The low reliability of synaptic transmission is at the heart of neural computation~\citep{Seung2003}. 
The brain not only overcomes such unreliability, but can even use it to drive learning~\citep{Kappel2015,Aitchison2021, Kasai2021}. 
Synaptic transmission is driven by dynamic processes among the molecules that make up the synapse, and probabilistic approaches are thought to be effective in understanding their robustness and plasticity~\citep{Choquet2013}. Furthermore, it has been reported that spine sizes exhibit dynamics arising from multiplicative updates~\citep{Loewenstein2011}.  
In general, multiplicative weight updates (MWU) attracted attention in various fields~\citep{Arora2012} such as machine learning~\citep{Littlestone1988,Jyrki1997}, optimization~\citep{Yoav1997}, game theory~\citep{Grigoriadis1995} and neuroscience~\citep{cornford2024brain}. 

For a loss function $\ell(\vparam)$, in contrast to gradient descent (GD), MWU modify the weights $\vparam$ via multiplication, for instance, consider the following versions:
\begin{equation}
    \text{GD: } \, \vparam \gets \vparam - \eta \nabla \ell(\vparam) \qquad \text{vs.} \qquad \text{MWU: } \, \vparam \gets \vparam \odot \exp(-\eta \nabla \ell(\vparam) \odot\operatorname{sign}(\vparam)).
    \label{eq:gdvsmwu}
\end{equation}
Here, $\odot$ is an element-wise product, $\operatorname{sign}(\vparam)$ is a vector of $-1$, $0$ and $1$ depending on the sign of the entries, $\eta$ is a learning rate, and exponentiation operations are performed element-wise.
\citet{Bernstein2020} applied the above MWU to ANNs and claim that an advantage over GD is that the size of the update is proportional to the magnitude of the weight, which they claim has stabilizing properties. However, their experiments have shown that MWU can overfit because the weights can grow exponentially. \citet{Bernstein2020} propose a method called Madam which clips the weights to reduce overfitting, but it is more desirable to handle this through regularization. Therefore, an effective regularization method for stabilizing learning in multiplicative weight updates is required.

\subsection{Multiplicative Noise Injection}

Biological synapses exhibit proportional changes in both activation-dependent plasticity and activation-independent spontaneous fluctuations, as discussed by~\citet{Loewenstein2011}. The former corresponds to multiplicative update dynamics. 
The latter spontaneous multiplicative fluctuations can be modeled as a log-normally distributed noise injection. This is in tandem with experimental findings which show that the spine size observed in biological synapses follow a log-normal distribution~\citep{Loewenstein2011}. There is no corresponding noise used in the Madam update~\citep{Bernstein2020}, but noisy fluctuations may play an important role in suppressing excessive weight growth. This motivates the use of noise injection in ANN training. 

Adding noise to the weights of ANNs during training also improves the generalization performance~\citep{an1996effects,orvieto2022anticorrelated}. Noise injection proportional to the magnitude of the weights has recently also shown impressive results for generalization~\citep{bisla2022low}, as well as for improved robustness to corruptions~\citep{Trung2024}. However, these works use normal distributions which are unnatural in a multiplicative setting. Instead, sampling from the log-normal distribution, %
\begin{wrapfigure}{r}{0.285\textwidth}
  \centering
  \vspace{-10pt}
  \includegraphics[width=0.25\textwidth]{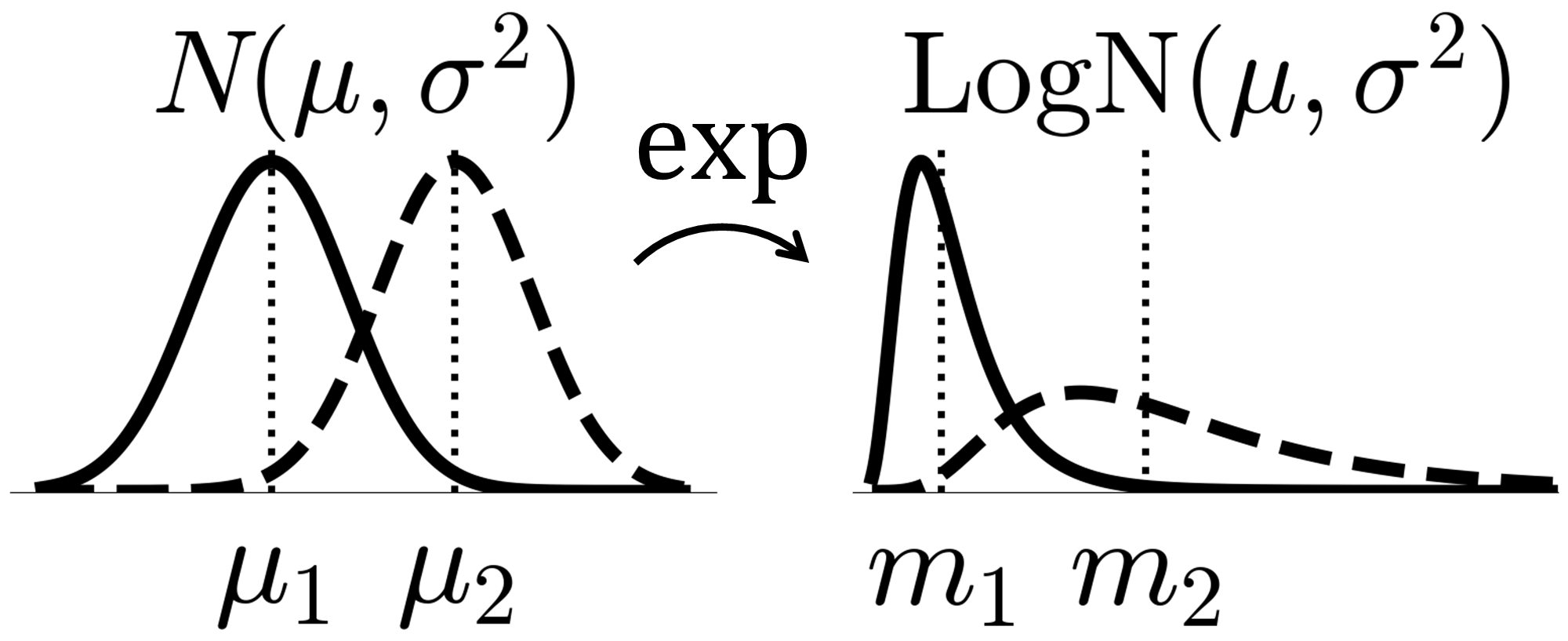}
  \vspace{-14pt}
  \label{fig:densitys}
\end{wrapfigure}
\begin{equation}
q(\theta)=\operatorname{LogN}(\theta \, | \, \mu,\sigma^2) :=  \frac{1}{\theta \sigma \sqrt{2\pi}} \exp\!\left( -\frac{(\log \theta - \mu)^2}{2\sigma^2} \right),
\label{eq:lognorm}
\end{equation}
is more natural. The figure shows a normal distribution $N(\mu,\sigma^2)$ and a log-normal distribution $\mathrm{LogN}(\mu,\sigma^2)$. Both use fixed $\sigma$ with two $\mu$ values. Here $m = \exp(\mu)$ is the median of the log-normal distribution.
Sampling from a log-normal incorporates proportional noise injection, since if $\varepsilon \sim \operatorname{LogN}(0, \sigma^2)$, then $m \varepsilon \sim \operatorname{LogN}(\log m, \sigma^2)$. The standard deviation of \(m\varepsilon\) is,
\begin{equation}
\mathrm{std}[m\varepsilon]= \sqrt{\mathrm{Var}[m\varepsilon]}
 = me^{\frac{\sigma^2}{2}} \sqrt{\bigl(e^{\sigma^2}-1\bigr)},
\label{eq:lognormstd}
\end{equation}
which is proportional to its mean \(\mathbb{E}[m\varepsilon]=m e^{\frac{\sigma^2}{2}}\) for fixed $\sigma$.
We propose to use the log-normal distribution as an approximate posterior in variational learning, which we introduce next. We will see that this yields more natural multiplicative dynamics with inherent multiplicative noise injection. 

\subsection{Variational Learning for Neural Networks}
Variational learning is formulated as an optimization problem over probability distributions $q(\vparam)$, 
\begin{equation}
\min_{q \in \mathcal{Q}} ~ \mathbb{E}_{\text{$\vparam$} \sim q}[\ell(\vparam)] + \tau \dkl{}{q(\vparam)}{p_0(\vparam)},  
\label{eq:variational}
\end{equation}
where $p_0(\vparam) \propto \exp(-R(\vparam))$ is a prior, \(\tau > 0\) is a temperature parameter and \smash{\(\ell(\vparam) = \sum_{i=1}^N \ell_i(\vparam)/N\)} denotes the empirical risk. For $\tau = 1$ and $\ell_i(\vparam) = -N\log p(\mathcal{D}_i \, | \, \vparam)$, this problem targets the Bayesian posterior~\citep{Zellner1988}, where \(\mathcal{D}_i\) denotes the i-th training example. $\mathcal{Q}$ is the set of candidate distributions, and by minimizing \eqref{eq:variational} one searches for the candidate in $\mathcal{Q}$ closest to the exact Bayesian posterior, where closeness is measured by the Kullback-Leibler divergence.
Variational learning for neural networks is typically implemented using stochastic gradient methods which leads to methods with (adaptive) noise injection~\citep{Graves2011, Khan2023}. 

In the past, multiplicative noise injection has been considered for Bayesian learning~\citep{Yeming2018} but is typically used with additive GD updates. Recently, multiplicative updates have been derived from a variational viewpoint by using distributions parametrized via Lie-groups~\citep{Kiral2023}. We use this framework to derive a new update because it naturally combine multiplicative updates with multiplicative noise injection. 
We use Algorithm~2 in \citet[App.~A.4]{Kiral2023} which optimizes Eq.~\ref{eq:variational} using the following updates:
\begin{equation}
\begin{aligned}
    \vg &\gets \vparam \odot(\nabla \ell(\vparam) + \tau \nabla R(\vparam)) - \tau, \, \text{ where }
        \, \vparam \gets \vm \odot \boldsymbol{\varepsilon} \, \text{ and } \, \boldsymbol{\varepsilon} \sim \operatorname{LogN}(\boldsymbol{\varepsilon} \, | \, 0, \sigma^2 \vI), \\
\vnu &\gets \beta \vnu + (1-\beta) \vg,  \\
\vm &\gets \vm \odot \exp\bigl(-\eta\, \vnu\bigr). \label{eq:lg-updates}
\end{aligned}
\end{equation}

Here, exponentiation operations are element-wise, 
$\eta>0$ is a learning rate and $\beta \in [0,1)$ a momentum parameter. From the viewpoint of Lie-groups, $\vg$ and $\vnu$ are tangent vectors and the $\exp$-function is \emph{the} natural mapping connecting the tangent space to the parameter space. We refer to \citet[Sec.~3.5]{Kiral2023} for details on the mathematical background and remark that the above method can also be viewed as gradient descent on the $\mu$-parameter of the log-normal distribution in Eq.~\ref{eq:lognorm}.

We specialized the updates in \citet[App.~A.4]{Kiral2023} to the case of $\mathcal{Q}$ being the set of log-normal distributions from Eq.~\ref{eq:lognorm} with fixed $\sigma^2\vI$ and update $\vmu$ through $\vm = \exp\vmu$, where $\vm \in \mathbb{R}_{>0}^P$ is the median of the log-normal distribution and belongs to the multiplicative Lie-group that parametrizes the distribution. In contrast to the widely used MWU in Eq.~\ref{eq:gdvsmwu}, the updates in Eq.~\ref{eq:lg-updates} (i) use noise injection, (ii) multiply $\nabla \ell (\vparam)$ with the weight $\vparam$ and (iii) incorporate an explicit regularizer $R(\vparam)$. 
Eq.~\eqref{eq:lg-updates} will be the starting point for our proposed method.

When training ANNs with MWUs in Eqs.~\ref{eq:gdvsmwu}~and~\ref{eq:lg-updates}, a caveat is that they do not change the sign of the weights. However, when weights can only take specific fixed signs, the expressive power of the network is limited~\citep[Section~8]{Bernstein2020}. \citet[Algorithm~2]{Kiral2023} and also~\citet{Jyrki1997,Ghai2020} adopt the “EG$\pm$ trick” which uses positive weights \(\vparam_+\) and negative (minus) weights \(\vparam_-\) for a single weight \smash{\(\vparam_{\text{trick}} = \vparam_+ - \vparam_-\)}. This form models the combined output of excitatory and inhibitory neurons,
see \cite[Figure 1c]{Kiral2023}. A biological interpretation of this learning rule only updating the magnitudes of weights respects Dale's law for ANNs~\citep{Amit1989}. We also adopt the same EG$\pm$ trick in our proposed method.

\subsection{Stable Low-Precision Training}

There are good reasons to consider low-precision data formats for ANNs as described by~\citet{Bernstein2020}, and we evaluate the performance of the proposed method when using low-precision data formats. From a biological point of view the spine size of a synapse can be distinguished into 26 levels, arranged logarithmically, and it is thought that each synapse has 4.7 bits of information capacity \citep{Bartol2015}.
This is substantially lower than the representational range of standard 32-bit floating-point formats, suggesting considerable potential to reduce memory footprints of ANNs.
ANNs have been frequently explored to utilize low-precision data formats~\citep{Matthieu2015,Kalamkar2019,Micikevicius2022}.
When combined with dedicated hardware accelerators, low-precision formats can substantially improve the throughput of matrix multiplications (GEMMs)~\citep{Paulius2018, Naigang2018, Houwen2023,Ruizhe2025}.
Furthermore, attempts have been made to significantly improve power efficiency through co-design approaches of data formats and dedicated hardware units~\citep{Edward2017, Kodai2018, Zhao2022, Haghi2024}.
\paragraph{\textcolor{black}{Microscaling (MX) Data Formats}}

Microscaling (MX) data formats~\citep{Rouhani2023OCP} is one family of the proposed formats in~\cite{Rouhani2023} with the aim of achieving wide dynamic range with limited bit width. In the MX data formats a collection of \(k_\text{mx}\) numbers is represented jointly, using very low precision private parts $p_\text{mx}$ and and a shared integer exponent \(s_\text{mx}\). In the standard MX settings, \(k_\text{mx} = 32\), and the shared scale variable \(s_\text{mx}\) is of type INT8. For each private element \(p_\text{mx}\) we represent the numbers \(2^{s_\text{mx}-b_\text{mx}}p_\text{mx}\) where \(b_\text{mx}\) is called a bias. For example in the MXFP6 and MXFP4 data formats the bias term is $1$ and the private elements $p_\text{mx}$ either use the FP6 (E2M3) representation with $1$-bit sign, $2$-bit exponent and $3$-bit mantissa, or use the FP4 (E2M1) representation with $1$-bit sign, $2$-bit exponent, and $1$-bit mantissa. MX data formats have also been considered for deployment of LLMs because it enables improved throughput of matrix multiplications~\citep{Bita2023,Verrilli2024,Tseng2025,Ramani2025}.

\paragraph{Rounding Error Issue in Low-Precision Deep Learning}

{~\citet{Bernstein2020} argues that multiplicative weight updates are effective for low-precision data representations, such as logarithmic formats (no mantissa) which emphasize dynamic range. ~\citet{Zhao2022} points out that for additive weight updates used in logarithmic formats~\citep{Daisuke2016}, the impact of rounding errors increases with weight value, which may lead to instability in learning. On the other hand, they explain that with multiplicative weight updates, the update becomes larger as the magnitude of value becomes larger, so the impact of rounding errors is reduced compared to additive updates. 
Rounding errors are roughly proportional to the magnitude of the value, a fact which is also true in floating point formats, and not just logarithmic formats.
In mixed-precision training~\citep{Paulius2018}, storing weights in 32-bit precision while performing matrix multiplications in a low-precision format (e.g. bfloat16) is standard. In fact, it has been reported that training performance in LLM training degrades due to rounding errors when weights are stored in bfloat16~\citep{Yu2024}. 

We aim to reduce the possibility that the effect of noise will be lost due to quantization, even when noise-injected weights are quantized to a low-precision data format.
To improve the throughput of matrix multiplications, we cannot mitigate the impact of rounding errors by using a high-precision data format, as has been done in some studies of weight updates.
In the case of multiplicative noise, weight perturbations scale with the magnitude of the weights and are less likely to be erased by the rounding inherent in low-precision quantization.

\paragraph{Stable Training by Proportional Fluctuation in Low-Precision Data Format}
Stochastic rounding is a technique for inducing fluctuations in the rounding process to stabilize training~\citep{Gupta2015}.}
One randomly rounds numbers to the nearest representable low-precision value such that the expected value of the rounded result is equal to the original number~\citep{Croci2022}. This has been shown to enable stable learning in small networks~\citep{Gupta2015,Zhang2022}. Even in LLMs, full-scratch training using MX data format~\citep{Rouhani2023OCP} with stochastic rounding in the backward pass tends to retain performance~\citep{Tseng2025}. Stochastic rounding has also been adopted for fine-tuning low bit-width 
model ensembles~\citep{Giung2024}. 

In the quantization step following our multiplicative noise injection, values may occasionally be rounded to points farther from their nearest-neighbor. 
Considering that nearest-neighbor rounding in quantization does not always yield optimal results~\citep{Nagel2020a},
our work demonstrates that the stochasticity introduced by this noise injection may stabilize learning in low-precision formats, possibly via a mechanism analogous to stochastic rounding.

\begin{algorithm}[!t]
  \caption{Proposed Log-Normal Multiplicative Dynamics (LMD) Optimizer}
  \label{alg:lmd}
  \begin{algorithmic}[1]
  \REQUIRE Learning rate $\eta > 0$, temperature $\tau>0$ , $\beta_1, \beta_2 \in [0, 1)$, 
  \STATE  Initialize $\vm = \left[\begin{smallmatrix} \vm_+ \\ \vm_-\end{smallmatrix}\right] \in \mathbb{R}^{2P}$ by Sec. \ref{sec:initialization},
  $\vnu = \left[\begin{smallmatrix} \vnu_+ \\ \vnu_-\end{smallmatrix}\right] = \mathbf{0} \in \mathbb{R}^{2P}$ and 
  $\vA^\top = \left[\begin{smallmatrix} \vI \\ \scalebox{0.75}[1.0]{$-$} \vI\end{smallmatrix}\right] \in\mathbb{R}^{2P \times P}$
  
  \WHILE{not converged}
  \STATE $\boldsymbol{\varepsilon} \sim \operatorname{LogN}(0,\sigma^2 \vI_{2P})$ \hfill \textcolor{gray}{\small{\# sample log-normal noise}}
  \STATE $\vparam \gets \vm \odot\boldsymbol{\varepsilon}$ where $\odot$ is element-wise product \hfill \textcolor{gray}{\small{\# multiplicative noise injection}}
  \STATE $\vg \gets \vparam \odot \vA^\top \hat \nabla \ell(\vparam_{\text{trick}})$ where $\vparam_{\text{trick}} = \vA \vparam$ \hfill \textcolor{gray}{\small{\# compute loss gradient using EG$\pm$ trick}}
  \STATE $\vr \gets \tau (\vparam \odot \nabla R(\vparam) - 1)$\hfill \textcolor{gray}{\small{\# compute regularization gradient}}
  \STATE $\vnu \gets \beta_2 \vnu +(1 - \beta_2) \vg$ and  $\vnu_\text{temp} \gets \beta_1 \vnu +(1 - \beta_1) \vg $\hfill \textcolor{gray}{\small{\# update momentum}}
  \STATE $\vm \gets$ $\vm \odot \exp\left(-\eta \left(\operatorname{sign}(\vnu_\text{temp}) +  \vr \right) \right)$\hfill \textcolor{gray}{\small{\# multiplicative update of median}}
  \ENDWHILE
  \end{algorithmic}
\end{algorithm}

\section{Log-Normal Multiplicative Dynamics for Training Large Networks}
We extend Eq. \ref{eq:lg-updates} to incorporate a design similar to synaptic multiplicative dynamics to enable stable low-precision training in large networks to overcome the small-scale limitation of \citet{Kiral2023}.

As ~\citet{Ilya2019} show, incorporating gradient and weight penalties with momentum accumulation tends to destabilize learning and degrade performance.
This instability arises because, when either the gradient or the penalty dominates, the accumulated moment neglects the other term~\citep{Bjorck2021}.
LMD addresses this issue by decoupling the gradient \(\vg\) and the regularizer \(\vr\) as in step 5 and 6 of Algorithm \ref{alg:lmd} as used in AdamW. We also use signed gradient momentum~\citep{Bernstein2018} and independent weight penalties, alongside two momentum coefficients, \(\beta_1\) and \(\beta_2\), as in Lion~\citep{Chen2023}. 
The default exponential moving average (EMA) coefficient for momentum \(\vnu\) is \(\beta_2=0.99\) in step 7 of Algorithm \ref{alg:lmd}. In each update, we interpolate between the current gradient $\vg$ and the momentum $\vnu$ using \(\beta_1=0.95\), and call that interpolation $\vnu_\text{temp}$. We then apply a sign operation to yield the update rule in step 8 of Algorithm \ref{alg:lmd}:
\begin{equation}\label{eq:lmd-updates}
  \begin{aligned}
    \quad \vm &\gets \vm \odot \exp\bigl(-\eta\, (\operatorname{sign}(\vnu_\text{temp}) + \vr)\bigr) .
  \end{aligned}
\end{equation}
We assume a separable component-wise regularizer $R(\vparam) = \sum_{i=1}^{2P} \tilde R(\param_i)$ and set \(\tau^{-1} := \tilde R'(1)-1\) as such that the individual entries of $\vr$ are \(r_i=1\) when the entries of $\vparam$ are \(\param_i = 1\). This corresponds to soft clipping for weights \(\vparam\) greater than 1 by combination of signed moments.
We only need to maintain two variables, \(\vm\) and \(\vnu\). Then, \(P\) weight parameters require additional \(P\) variables. EG\(\pm\) trick doubles the weights, so LMD holds \(4P\) parameters in total. By contrast, AdamW keeps \(P\) weights plus \(2P\) momentum terms for a total of \(3P\), meaning LMD uses one additional \(P\)-dimensional vector compared to AdamW.

LMD uses Monte Carlo (MC) sampling and can efficiently perform multi-MC samples which effectively reduce the variance \citep{Kingma2015}, in multi-GPU environments.
When using such multi-MC sampling, \(\vg\) and \(\vr\) as in step 5 and 6 of Algorithm \ref{alg:lmd} are redefined as:
\begin{equation}\label{eq:multi-mc}
  \begin{aligned}
    \vg &\gets \frac{\sum_{j,s}
    \vparam_j^{(s)} \odot \vA^\top \hat \nabla \ell(\vA \vparam_j^{(s)})}{J \cdot S},
    \, \vr \gets \frac{\sum_{j,s} \tau (\vparam_j^{(s)} \odot \nabla R(\vparam_j^{(s)}) - 1)}{J \cdot S}, \\
  \end{aligned}
\end{equation}
where \(J\) is the number of devices, and \(S\) is the number of MC samples per device. We use a different random sample \smash{\(\vparam_j^{(s)}\)} on each device \(j\) and for each MC sample \(s\). 
LMD is implemented as a drop-in replacement for Adam, similar to IVON~\citep{Shen2024}. For details, see Appendix~\ref{app:settingdetails}.

\subsection{Multiplicative Weight Decay}
In LMD we perform weight decay by assuming a log-normal distributed prior in Eq.\ref{eq:variational}.
We choose a component-wise log-normal prior \smash{$\tilde R(\theta) = -\log p_0(\theta) = -\log (\operatorname{LogN}(\theta \, | \, \log m_r,\sigma^2))$}, which leads to the following regularizer gradient $\vr$ used in Eq.\ (\ref{eq:lmd-updates}), here written for a single component:
\begin{equation}\label{eq:lmd-decay}
  r  = \tau (\theta \tilde R'(\theta) - 1) =  \tau\Bigl[\theta\,\underbrace{\frac{1}{\theta} \left( 1 + \frac{\log \theta - \log m_r}{\sigma_p^2} \right)}_{\tilde R'(\theta)} - 1 \Bigr] = \tau\left(\frac{\log \theta - \log m_r}{\sigma_p^2}\right).
\end{equation}
To understand the effect of the choice of the log-normal prior on the learning dynamics, we take the expectation of the regularizer gradient $r$ multiplied with the learning rate \(\eta\). This leads to, \(\mathbb{E}[ -\eta \,r] = \mathbb{E}[-\alpha \log( \frac{\theta}{m_{\text{r}}})] = \log ( (\frac{m}{m_r})^{-\alpha} )\) where \(\alpha = \tfrac{\eta\gamma}{\sigma^2}\geq 0\) and where we used that \(\log \theta\) is distributed normally with mean \(\log m\). Using these, the LMD update can be expressed as:
\begin{align}
  \vm &\leftarrow (\vm^{1-\alpha}\,m_{\mathrm r}^{\alpha})\,\odot
             \exp\bigl(-\eta\,\operatorname{sign}(\vnu_\text{temp})\bigr),
     \label{eq:theta_update_pow}\\
     \Leftrightarrow \log \vm
     &\leftarrow (1-\alpha)\,\log \vm + \alpha\,\log m_{\mathrm r}
      - \eta\,\operatorname{sign}(\vnu_\text{temp}).
     \label{eq:theta_update_log}
\end{align}
As can be seen from Eq.\ref{eq:theta_update_log}, this simply corresponds to a weight decay in logarithmic space.
Note that unlike usual weight decay, this penalty does not force weights to zero. Instead, it makes \(m\) gravitate towards \(m_r\). When used in conjunction with the plus-minus trick, then positive and negative weights move to \(m_r\) and the expected value of \(\vparam_+ - \vparam_-\) is zero. 
Larger values of \(m_r\) lead to larger multiplicative noise injection but we still have zero-mean in expectation. 
As shown in  \citet{Trung2024}, multiplicative weight noise is thought to mimic activation fluctuations as follows:
\begin{equation}
  z_j=\textstyle\sum_i \bigl(m_{ji}(1+\sigma\varepsilon_{ji})\bigr)x_i = \textstyle\sum_i m_{ji}\bigl(\underbrace{(1+\sigma\varepsilon_{ji})x_i}_{\text{perturbed activation}}\bigr),
\end{equation}
where \(x_i\) is the input (activation), \(z_j\) is the \(j\)-th pre-activated neuron and \(\varepsilon_{ji} \sim N(0,1)\) is the injected noise. Therefore, both weights that do not contribute to inference and remain near \(m_r\) can be thought of as encouraging the emulation of activations rather than discarding them as in zero (pruned) weights.

\subsection{Initialization}
\label{sec:initialization}
We also adjust the initialization to facilitate parameter exploration in the vicinity of~\(m_r\).  Denote by \(\theta_0\) the default initiailization of the neural network, and define the entires of $\vm$ as:
\begin{equation}\label{eq:lad-init}
  \begin{aligned}
    m_{+,0}
    &=
    \begin{cases}
    \theta_0\exp\bigl(-\tfrac{\sigma^2}{2}\bigr)+m_r,
    & \theta_0>0,\\
    m_r,
    & \theta_0\le0,
    \end{cases}, \quad
    m_{-,0}
    &=
    \begin{cases}
    m_r,
    & \theta_0 >0,\\
    -\theta_0\exp\bigl(-\tfrac{\sigma^2}{2}\bigr)+m_r,
    & \theta_0\le0.
    \end{cases}
  \end{aligned}
\end{equation}
This ensures that the mean of the values of \(\vA \vparam\) in line 5 of Algorithm~\ref{alg:lmd} is \(\theta_0\).
Moreover, this allows us to reuse default initialization schemes which were devised for gradient-descent learning. 
For any parameters initialized to one, for instance, in scale parameters in normalization layers, we instead set
\smash{\(m_{+,0} = \exp(-\sigma^2/2)\)}, \(m_{-,0} = 0\), \smash{\(m_r  = \exp(-\sigma^2/2)\)}, \smash{\(\tau^{-1}  = 2 \tilde R'(2)-1\)}. That is, the scale parameter is set so that the mean is 1 and the maximum is soft clipped to 2, which is heuristic but found to work well in practice and to stabilize the training.

\section{Experiments}
\paragraph{Microscaling (MX) Data Format Settings}
In this study, both MXFP6 and MXFP4 formats are used as default settings.
To enable training of large networks on existing GPUs, we use a custom CUDA library to emulate the MX data format~\citep{Microscaling2023, Bita2023}. Under this emulation, low-precision matrix multiplications are performed in MXFP6 or MXFP4. In our experiments, vector operations such as activations and all backward-pass computations are carried out in bfloat16. We note that the internal state variables of the optimizer still remains in FP32.

\paragraph{ViT Experimental Settings}
We employed a Vision Transformer (ViT) with an embedding dimension of \(384\), \(6\) attention heads, and \(12\), transformer layers. Input images were resized to \( 224 \times 224\) pixels and split into \(16 \times 16\) patches. Instead of using the [CLS] token embedding, we applied global average pooling over patch embeddings for classification. Positional embeddings were treated as learnable parameters. ViT adopts the implementation of PyTorch Image Models~\citep{Wightman2019}.
All ViT experiments were run on the ImageNet dataset~\citep{Deng2009} for 200 epochs with a batch size of 4,096. We applied standard data augmentation, random resized crop to \( 224 \times 224\) and random horizontal flip. Training was performed on eight NVIDIA H100 GPUs (96 GB each), requiring about 6 to 8 hours; when using the MX format emulator extended this to roughly half a day.
we distributed one GPU per node across eight nodes.
In LMD experiments, each GPU samples weights from a different random seed, resulting in MC sampling beging \(J = 8\) and  \(S = 1\). The learning rate follows a cosine-annealing schedule down to zero, with a linear warmup over the first 10,000 steps. 
For LMD, we set the learning rates to \(\eta=0.005,\sigma = 0.125 \), and \(m_r = 0.01 \times \exp\bigl(\tfrac{\sigma^2}{2}\bigr)\)\footnote[1]{\(m_r = 0.01\) is expected to give the same results.}. LMD did not employ gradient-norm clipping. For AdamW, we used a learning rate of \(0.001\), with \(\beta_{1}=0.9\), \(\beta_{2}=0.999\), and a weight decay of \(0.1\). For Madam, the learning rate was \(0.01\), and parameter values were clipped to a maximum magnitude of \(1\). Both AdamW and Madam employed gradient-norm clipping at \(1\).

\paragraph{GPT-2 Experimental Settings}
We trained the GPT-2 model \cite{Radford2019} on the OpenWebText dataset \cite{Gokaslan2019} for a total of \(25{,}000\) steps, processing approximately \(2.1\times10^{6}\) tokens per step (total \(\approx 52\) Billion  tokens) using \(32\) gradient accumulations. 
We used the nanoGPT framework~\citep{Karpathy2022} to train the model.
Training was performed on eight NVIDIA H100 GPUs (96 GB each), requiring about 6 to 8 hours; when using the MX format emulator extended this to roughly half a day but employed sixteen H100 GPUs.
We assigned one GPU to each node and determined the number of nodes based on the number of GPUs to be used.
In the LMD experiments, on each GPU, we sampled weights from a different random seed, and the number of MC samples per weight update was equal to the number of gradient accumulation steps with \(32\) times, i.e., \(J = 8\) and  \(S = 4\) were used. The learning rate decays to one-tenth of its initial value via cosine annealing, following a linear warmup over the first $2{,}000$ steps. We trained all GPT-2 models without bias parameters with 124 M parameters in total. For LMD, we set the initial learning rate to \(\eta=0.005,\sigma = 0.125 \), and \(m_r = 0.01 \times \exp\bigl(\tfrac{\sigma^2}{2}\bigr)\) (same as ViT settings). LMD employed gradient-norm clipping at \(10\). For AdamW, we used a learning rates of \(0.0006\), \(\beta_{1}=0.9\), \(\beta_{2}=0.95\), and a weight decay of \(0.1\). For Madam, the learning rate was \(1\times10^{-2}\) with a weight decay of \(0.1\).

\begin{figure}[!t]
  \centering
  \includegraphics[width=0.49\textwidth]{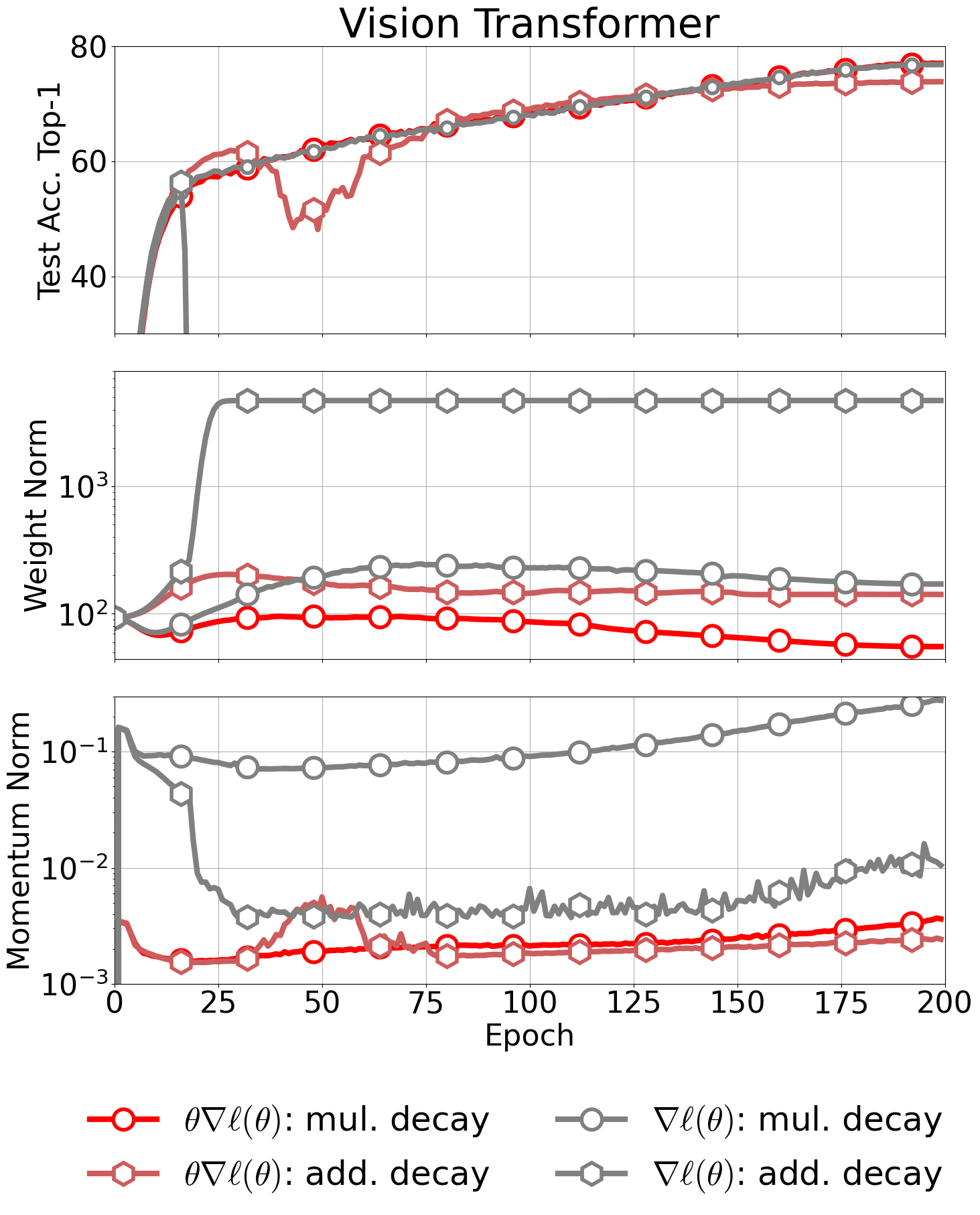}
  \includegraphics[width=0.49\textwidth]{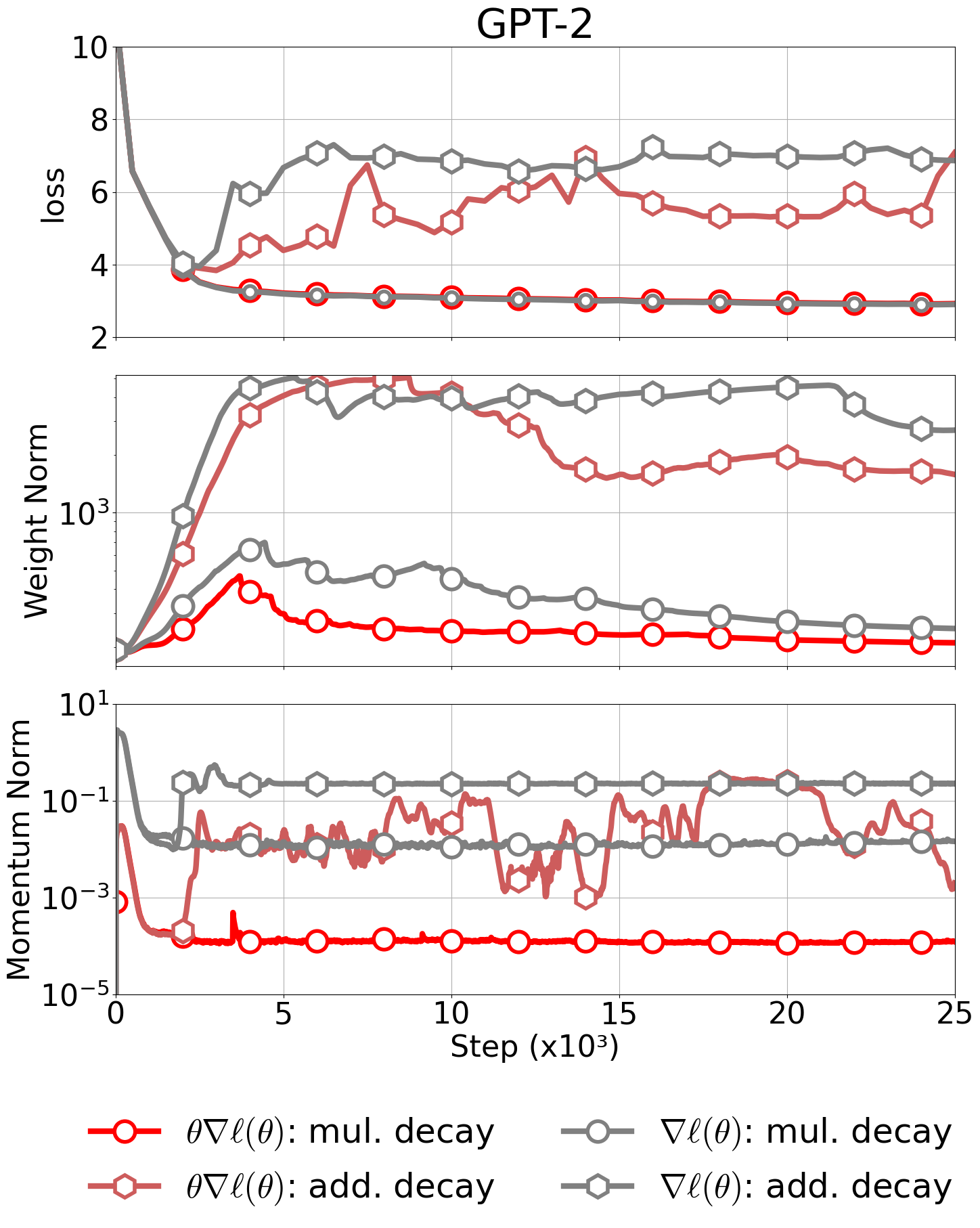}
  \caption{These plots show the effectiveness of multiplicative regularization in LMD. By scaling the gradient according to the weight magnitude and employing multiplicative weight decay in LMD, we observed consistently stable training dynamics in both ViT and GPT-2.}
  \label{fig:fig3}
\end{figure}

\subsection{Comparison with Existing Optimizers}
We compared LMD against AdamW and Madam (Figure \ref{fig:fig1}, \ref{fig:fig2}). The detailed results are shown in Table \ref{tab:results}. In ViT training, LMD was shown to overwhelmingly outperform both AdamW and Madam. In GPT-2 training, when the maximum sequence length (corresponding to “seq.\ len” in the Table \ref{tab:results}) was 4096, LMD achieved higher performance than any of the AdamW variants. In MXFP6 forward-pass training, AdamW tended to suffer degraded performance, whereas LMD exhibited no such degradation. This demonstrates that LMD can leverage low-precision forward computations yet still train both stably and with high accuracy. Madam performed worse in all settings, and as shown in Figure \ref{fig:fig2} its weight \(\ell_{2}\) norm tended to grow relative to the other optimizers, indicating unstable training. LMD, in contrast, nearly converged to the initial weight \(\ell_{2}\) norm, suggesting that it seeks optimal solutions in regions of parameter space where the weight norm remains small. The weight norms for LMD were computed by using the mean value of \(\vparam_+ - \vparam_-\).

\subsection{Effectiveness of Multiplicative Regularization in LMD}
We perform an ablation study to understand why LMD trains effectively. The regularization in LMD stems not only from the multiplicative weight update in Eq.\  (\ref{eq:gdvsmwu}) but also from scaling the gradient by the weight, as shown in Eq.\  (\ref{eq:lg-updates}). We show that these two features suppress the excessive weight growth characteristic of multiplicative updates. Figure \ref{fig:fig3} depicts a total of four cases: multiplicative weight decay or additive weight decay, each combined with either gradient scaling by weights or no scaling. Empirically, since the momentum \(\vnu\) does not change the trend based on the sign, we plot the \(\ell_{2}\) norm of the momentum for positive weights. The momentum norm captures the dynamics of the gradient magnitudes through accumulation of past gradients.
Additive weight decay was evaluated with \(r = (\theta - m_{\text{r}})/(1-m_{\text{r}})\). This means that when \(\theta\) is one, \(r\) also one.
Overall, multiplicative weight decay contributes more strongly to stabilizing LMD training compared to additive decay. Under additive decay, the weight \(\ell_{2}\) norm tends to increase, indicating weak regularization, and this effect is especially pronounced in GPT-2. Under multiplicative decay, when the gradient is scaled by the weight, fluctuations in the weight \(\ell_{2}\) norm are more gradual than in the unscaled case , clearly demonstrating effective regularization. The momentum norm is also visibly smaller in magnitude, indicating suppressed gradient variance.

\subsection{Effectiveness of Multiplicative Noise Injection for Low-Precision Training}
\begin{figure}[!t]
  \centering
  \includegraphics[width=0.49\textwidth]{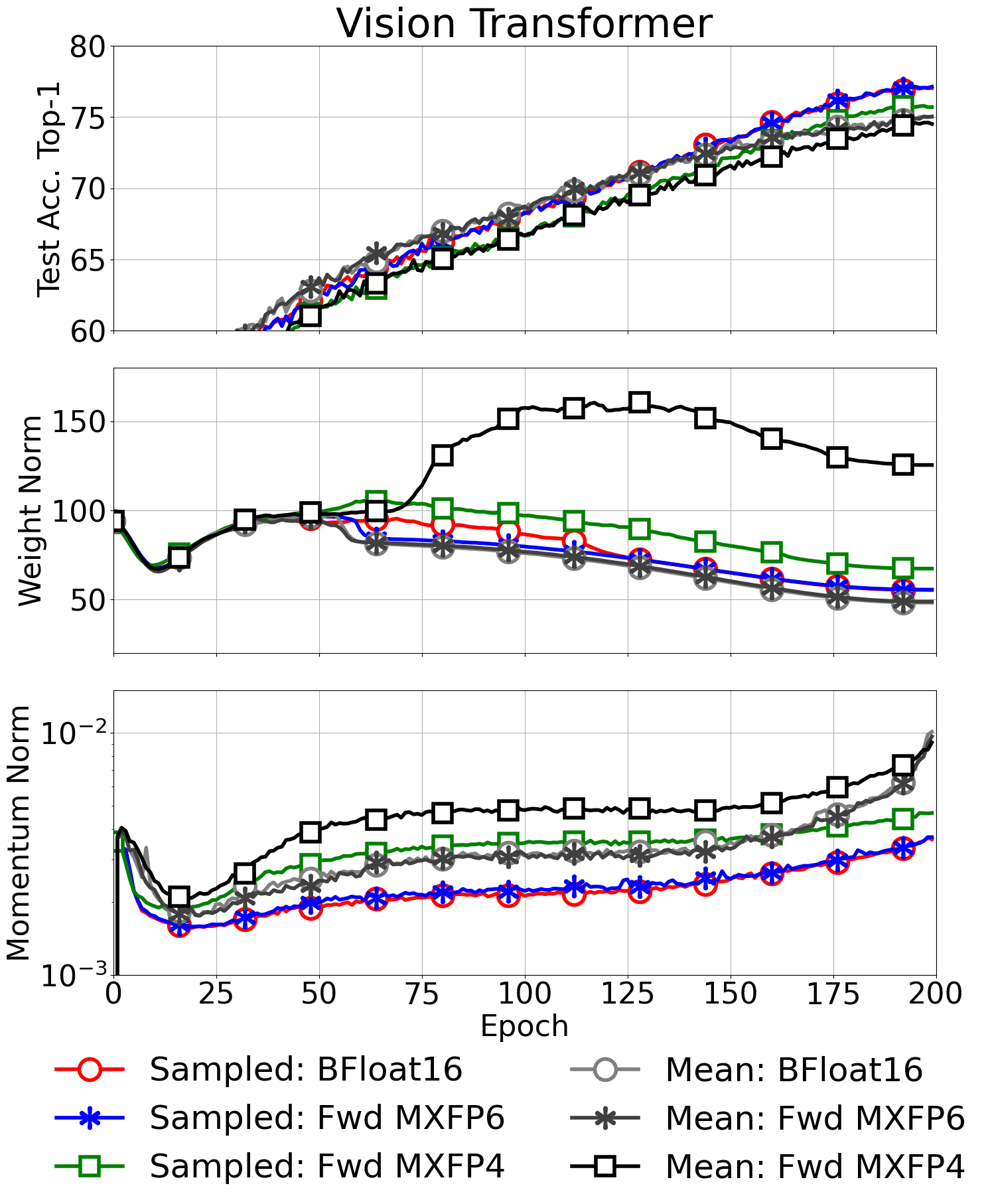}
  \includegraphics[width=0.49\textwidth]{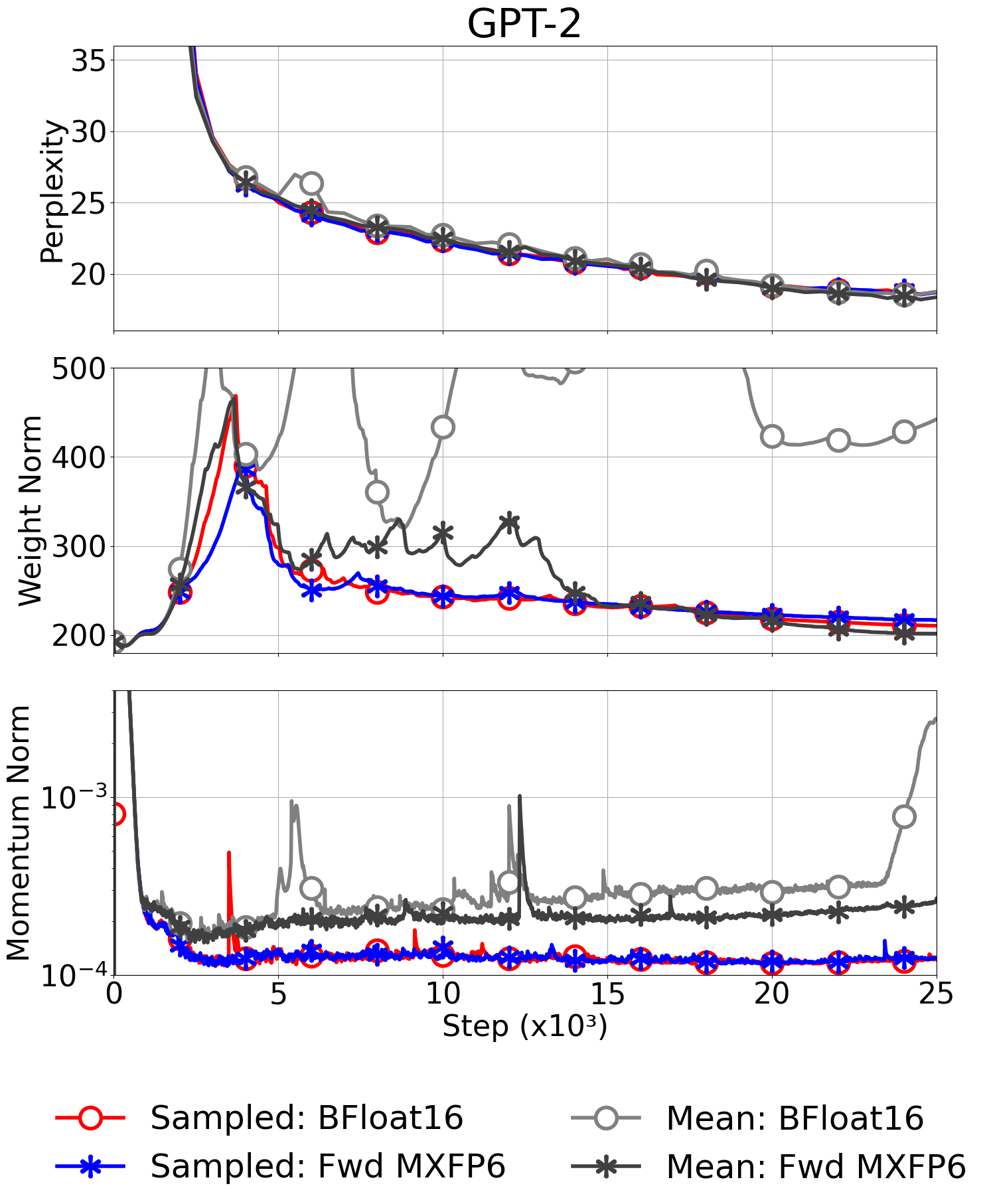}
  \caption{This panel shows the effectiveness of multiplicative noise injection for low-precision training. By sampling weights, it is possible to perform weight regularization and gradient stabilization.}
  \label{fig:fig4}
\end{figure}

We investigated the importance of multiplicative noise injection in low-precision training to use MX data format emulation. Multiplicative weight noise suppresses the growth of the the momentum norm and yields consistent learning dynamics regardless of whether a MXFP6 forward pass is used. Figure ~\ref{fig:fig4} plots four cases in GPT-2: training with bfloat16 forward passes and MXFP6 forward passes, each combined with either sampled training or mean training. In the case of ViT, six patterns are shown by adding MXFP4 results.
In the ViT experiments, sampled training clearly achieves higher accuracy than using only the mean training. This is because in the final phase of training (when cosine annealing has a strong effect), sampling better suppresses the growth of the \(\ell_{2}\) norm of the momentum, which avoid overfitting. Looking at the MXFP4 results, the mean training (black) shows a noticeably larger \(\ell_{2}\) norm of weights compared to the sampled training (green), indicating that sampling is an important regularizer in low-precision training. In the GPT-2 experiments, sampled training showed almost identical dynamics regardless of precision, demonstrating that LMD's multiplicative noise injection enables stable low-precision training. Conversely, mean training by MXFP6 showed a clear increase in \(\ell_{2}\) norm of weights, indicating insufficient regularization. Nevertheless, the final performance of the mean training using MXFP6 was close to that of sampled training. This suggests that the numerical error in MXFP6 may have played a role in regularization, but the dynamics are inconsistent with the bfloat16 result.

\newpage
\section{Discussion}
\paragraph{Limitations}
Despite the encouraging results, our study has several limitations. First, although we demonstrated stable training when using MXFP6 for forward passes, we have not evaluated low‐precision backward propagation, and because all experiments were conducted via emulation, we did not measure any real‐world speedups. Instead, our primary focus was to confirm whether the multiplicative update rule remains effective under operations subject to large numerical errors. 
Second, The EG\(\pm\) trick doubles the number of weight parameters that need to be learned, which may increase memory usage. However, in large transformers such as GPT~\citep{Vaswani2017}, the memory usage bottleneck lies not in the weight parameters themselves but in the activations \cite[Figure 2]{Chitsaz2024}. While the EG\(\pm\) trick doubles the number of weights, the number of weights used in matrix multiplication remains unchanged, so it does not affect the memory usage of the activations and computational costs of GEMMs. But, LMD requires one additional state vector compared to AdamW.
Finally, because we concentrated on optimizer stability when training from scratch, we have not examined the effectiveness of LMD for fine‐tuning or other downstream tasks.

\paragraph{Conclusions}
We proposed LMD, a multiplicative weight update (MWU) algorithm that incorporates the log-normal multiplicative dynamics observed in spines of biological synapses. The multiplicative regularization introduced by LMD suppresses the excessive weight growth that was a challenge in MWU, making it possible to train large-scale neural networks such as Vision Transformers (ViT) and GPT-2. Notably, even when using the low-precision MXFP6 data format for the forward pass, no performance degradation was observed. These results suggest that LMD has the potential to serve as an effective optimizer when utilized with energy-efficient hardware architectures designed for low-precision computations. 

\section*{Acknowledgments}
This work was supported by JST, CREST Grant Number JPMJCR2112 and the RIKEN Special Postdoctoral Researcher Program.
Computational resources were provided by the TSUBAME4.0 supercomputer at the Institute of Science Tokyo. 
We thank Bai Cong for setting up support for the GPT-2 experiments, and Christopher J. Anders for feedback on a draft of this paper.

%
%
%
%
%
%
\appendix
\section{Setting Details for the LMD Optimizer} \label{app:settingdetails}

\paragraph{Pytorch Implementation}
LMD is easy to use because it is implemented as a pytorch optimizer, like AdamW and Madam, and its implementation is very similar to IVON~\citep{Shen2024}\footnote{https://github.com/team-approx-bayes/ivon}, an optimizer for Bayesian deep learning. The simplest way to use it is shown in the Figure ~\ref{lst:lmd-train}.
\noindent

\begin{figure}[H]               
  \centering
\begin{lstlisting}
 import torch
+from lmd.lmd import LMD

 train_loader = torch.utils.data.DataLoader(train_dataset) 
 test_loader  = torch.utils.data.DataLoader(test_dataset) 
 model        = MLP()

-optimizer = torch.optim.AdamW(model.parameters())
+optimizer = LMD(model)

 for X, y in train_loader:

+  for _ in range(train_samples):
+    with optimizer.sampled_params():
       optimizer.zero_grad()
       logit = model(X)
       loss  = torch.nn.CrossEntropyLoss(logit, y)
       loss.backward()

     optimizer.step()
\end{lstlisting}
\caption{The simplest implementation of LMD in pytorch.}
  \label{lst:lmd-train}
\end{figure}

\section{Experimental Settings}

\paragraph{Madam Settings}
As discussed by ~\citet{Bernstein2020}, the performance of Madam varies significantly depending on the threshold of the weight clipping. Therefore, in this paper, we clip the magnitude of Madam's weights so that they do not exceed \(1\), consistent with the soft clipping applied to the weights of LMD. This is because, in Madam's original clipping settings, the initial \(\ell_{2}\) norm of the weights drops dramatically in the first update regardless of the gradient, making comparisons of training dynamics with LMD or AdamW impossible.

\paragraph{MX Data Format Emulation}
We applied the MX PyTorch Emulation Library\footnote{https://github.com/microsoft/microxcaling/tree/v1.1.0} for forward-only, low-precision experiments. In this emulation (see Figure \ref{fig:mx-emu}), activations and weights are first converted to the specified MX data format and then dequantized to bfloat16. Matrix multiplication is performed on these dequantized values; the result is then cast to bfloat16, after which activation and normalization layers are applied. Backward propagation is carried out in bfloat16.

\setcounter{figure}{5}
\begin{figure}
    \centering
    \includegraphics[width=0.9\linewidth]{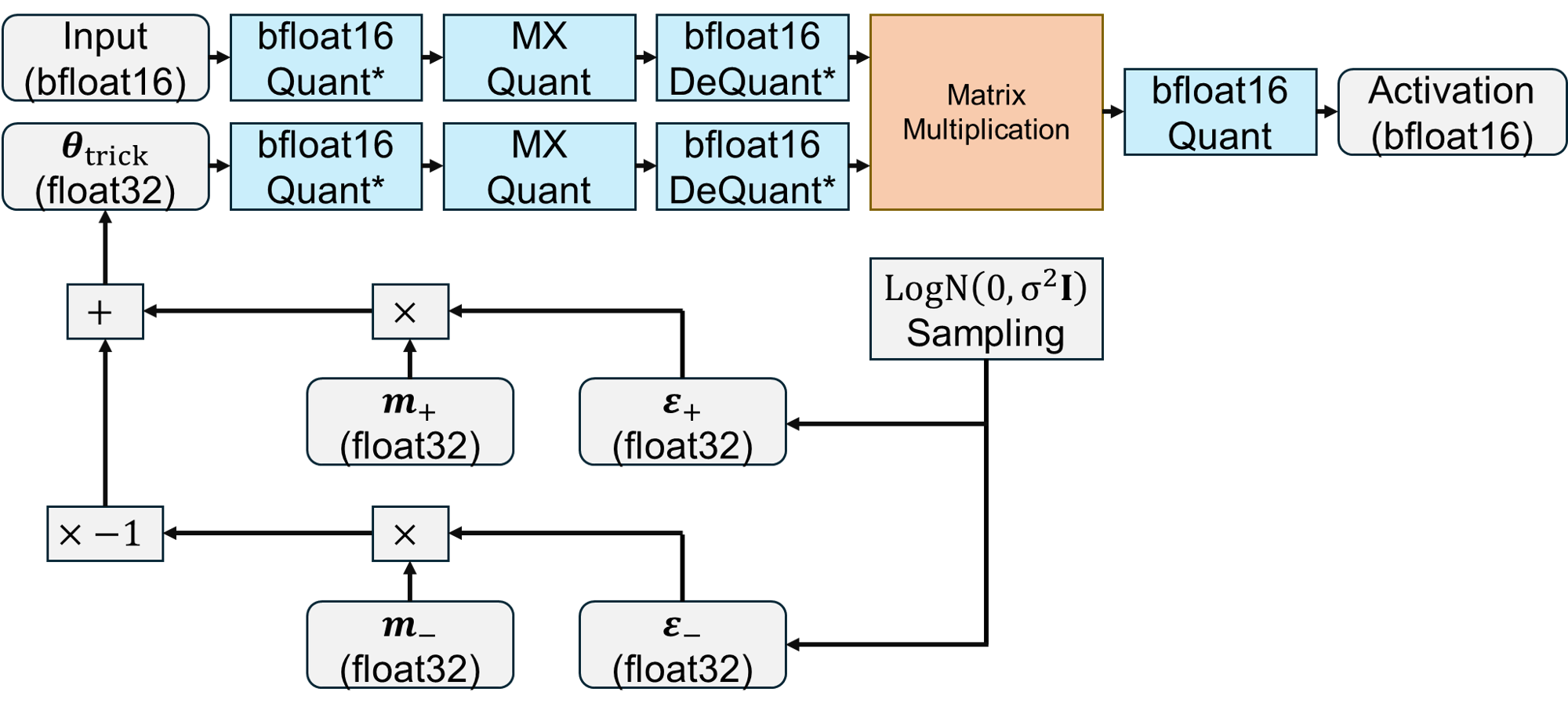}
    \caption{Overview of MX Pytorch Emulation for forward-pass with noise injection. Quantization (Quant) and Dequantization (DeQuant) of bfloat16 and quantization of the specified MX format are performed. The weights $\vm_+$ and $\vm_-$, as well as the sampled noises $\boldsymbol{\varepsilon}_+$ and $\boldsymbol{\varepsilon}_-$, were represented using float32. If you do not use emulation, the steps marked with * will be skipped. }
    \label{fig:mx-emu}
\end{figure}
\newpage
\section{Author Contributions Statement}
Authors list: Keigo Nishida (KN), Eren Mehmet Kıral (EMK), Kenichi Bannai (KB), Mohammad Emtiyaz Khan (MEK), Thomas Möllenhoff (TM)

KN conceived the concept of log-normal weight dynamics, underscored its strong connection to both neuroscience and computer-science approaches to machine learning, and led the project. KN also developed the core algorithms in Section 3, conducted all experiments with some advice from TM, and drafted the manuscript with assistance from EMK, MEK, and TM. EMK and KB refined the mathematical description; MEK and TM's variational learning insights shaped the method's direction; and EMK gave substantial feedback for the whole paper. The final manuscript was reviewed and approved by all authors.
\end{document}